\documentclass[11pt,a4paper]{article}
\usepackage[hyperref]{emnlp2020}
\usepackage{times}
\usepackage{latexsym}

\usepackage{microtype}
\usepackage{graphicx}
\usepackage{subcaption}
\usepackage{array}
\usepackage{lettrine}

\usepackage{amsmath}
\usepackage{booktabs}
\usepackage{svg}

\aclfinalcopy 

\title{FastFormers: Highly Efficient Transformer Models \\ for Natural Language Understanding}

\author{Young Jin Kim \\
  Microsoft \\
  One Microsoft Way \\
  Redmond, WA 98052, USA \\
  \texttt{youki@microsoft.com} \\\And
  Hany Hassan Awadalla \\
  Microsoft \\
  One Microsoft Way \\
  Redmond, WA 98052, USA \\
  \texttt{hanyh@microsoft.com} \\}

\date{}

\begin{document}
\maketitle
\begin{abstract}
Transformer-based  models are the state-of-the-art for Natural Language Understanding (NLU) applications. Models are getting bigger and better on various tasks. However, Transformer models  remain  computationally challenging  since they are not efficient at inference-time compared to traditional approaches. In this paper, we present \textit{FastFormers}, a set of recipes to achieve efficient inference-time performance for Transformer-based models on various NLU tasks. We show how carefully utilizing knowledge distillation, structured pruning and numerical optimization can lead to drastic improvements on inference efficiency. We provide effective recipes that can guide practitioners to choose the best  settings for various NLU tasks and pretrained models. Applying the proposed recipes  to the SuperGLUE benchmark, we achieve from 9.8x up to 233.9x speed-up compared to  out-of-the-box models on CPU. On GPU, we also achieve up to 12.4x speed-up with the presented methods. We show that \textit{FastFormers} can drastically reduce cost of serving 100 million requests from 4,223 USD to just 18 USD on an \textit{Azure F16s\_v2}\footnote{\url{https://docs.microsoft.com/en-us/azure/virtual-machines/fsv2-series}} instance. This translates  to a sustainable runtime  by reducing energy consumption 6.9x - 125.8x  according to the metrics used in the SustaiNLP 2020 shared task.

\end{abstract}

\section{Introduction}
Since \textit{BERT} \citep{devlin2018bert} has been introduced, Transformer \citep{vaswani2017attention}-based pretrained language models have dominated the  Natural Language Understanding (NLU) field. Transformer models have  provided unprecedented accuracy improvement compared to traditional models \citep{devlin2018bert, liu2019roberta}. However, the models' computational cost at inference time is prohibitively challenging to be widely adopted in real world production scenarios which requires low latency, fast inference  and low serving costs. In this work, we  present \textit{FastFormers}, a set of methods and recipes  that provides  highly efficient inference  for Transformer models which enables deployment in large scale  production scenarios. We specifically focus on the inference time efficiency since  it mostly dominates the cost of production deployment.

Mainly, we utilize three methods: Knowledge Distillation, Structured Pruning and Model Quantization. First, we investigate the efficacy  of various  Knowledge Distillation techniques to significantly reduce the size of the models with respect to the depth and hidden state sizes while preserving the accuracy. Second, we explore Structured Pruning that further reduces the size of the models by reducing the number of self-attention heads and the number of intermediate hidden states in the feed-forward layers  to  achieve more efficiency  while trying to preserve the accuracy as well. Finally, we explore Model Quantization which enables faster model executions by optimally utilizing hardware acceleration capabilities. On CPU, 8-bit integer quantization method is applied to utilize the most efficient CPU instructions available: \textit{Vector Neural Network Instructions (VNNI)}. On GPU, all the model parameters are converted into 16-bit floating point data type to maximally utilize efficient \textit{Tensor Core}s. Furthermore, computational graph optimizations which fuse multiple graph nodes are performed by utilizing \textit{onnxruntime}\footnote{\url{https://github.com/microsoft/onnxruntime}} library. In addition,  we explore optimal settings  of allocating CPU and GPU resources to achieve better utilization.

The proposed methods are optimized and evaluated on both  CPUs and GPUs which are the  most commonly available hardware platforms. Performance evaluations are conducted on SuperGLUE \cite{wang2019superglue} which is one of the general purpose open domain NLU benchmarks. For the efficiency measurement, wall clock times and energy efficiency are measured while performing inferences on the test sets of \textit{BoolQ}, \textit{CB}, \textit{COPA}, \textit{MultiRC}, \textit{ReCoRD}, \textit{RTE} and \textit{WiC} tasks from SuperGLUE. The energy efficiency is measured by an open source python library called \textit{experiment-impact-tracker} proposed in \cite{henderson2020towards}. To make sure the optimized models preserve similar accuracy, the accuracy of all the models is measured together.

The contributions of this paper are presenting a set of recipes for efficient inference of Transformer NLU models, analyzing  the effect of various optimization techniques and finally  making the code  publicly available~\footnote{\url{https://github.com/microsoft/fastformers}} to facilitate  utilizing \textit{FastFormers} for  efficient inference of Transformers models.

The rest of the  paper is organized as follows: Section~\ref{sec:kd} presents  Knowledge Distillation techniques, Section~\ref{sec:sp} presents Structured Pruning techniques, Section~\ref{sec:qn} discusses Model Quantization approaches, Section ~\ref{sec:rt} presents runtime optimization techniques, Section~\ref{sec:results} presents results on various tasks and finally Section~\ref{sec:discuss} concludes the findings and future directions.

\section{Knowledge Distillation}
\label{sec:kd} 
Knowledge distillation \cite{hinton-kd-2015}  is a well known model compression technique where the large model is used as a teacher for a smaller 
student model. The knowledge distillation is the process of training the student model to mimic the behaviour of the larger teacher model. Knowledge distillation has been shown to improve  the efficiency of the Transformer-based architectures for the NLU tasks \citep{sanh2019distilbert, jiao2019tinybert} as well as natural language generation tasks such as machine translation \cite{kim2019research}. 

\begin{figure*}[hbt!]
\begin{subfigure}{.5\textwidth}
  \centering
  \includegraphics[width=.8\linewidth]{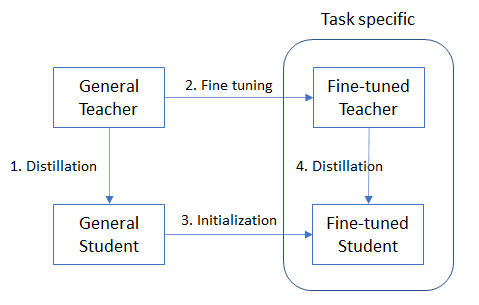}
  \caption{Task specific distillation to general distill models}
\end{subfigure}%
\begin{subfigure}{.5\textwidth}
  \centering
  \includegraphics[width=.8\linewidth]{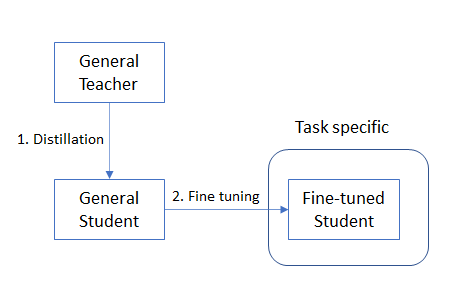}
  \caption{Fine-tuning of general distilled models}
\end{subfigure}
\caption{Knowledge distillation methods}
\label{fig:distillation}
\end{figure*}

\paragraph{Knowledge distillation methods:}
We utilize two different distillation approaches, namely \textit{task-specific} and \textit{task-agnostic} distillation. In the task-specific distillation, we distill  fine-tuned teacher models into smaller student architectures following the procedure proposed by \textit{TinyBERT} \citep{jiao2019tinybert}. In the task-agnostic distillation approach, we directly apply fine-tuning on general distilled models to tune for a specific task. These two methods are illustrated in Figure \ref{fig:distillation}. In the illustration, the preceding number attached to each arrow means the order of distillation steps.  We use soft cross-entropy function as the knowledge distillation loss function in all our experiments as used in \citep{sanh2019distilbert, jiao2019tinybert}.

As teacher models, we choose 12 stacked layer \textit{BERT}\citep{devlin2018bert} and \textit{RoBERTa}\citep{liu2019roberta} models with 768 hidden state dimension and 12 self-attention heads. This is referred as \textit{Base} size from the original BERT paper. In particular, we use HuggingFace's pretrained BERT and RoBERTa \footnote{\url{https://github.com/huggingface/Transformers}} models.

In \textit{task-specific} scenario, we observe that the initialization of the student models affects the final accuracy of the distilled models. We utilize open domain pre-distilled models, namely  \textit{distilroberta-base}\footnote{\url{https://huggingface.co/distilroberta-base}} and \textit{TinyBERT}\footnote{\url{https://github.com/huawei-noah/Pretrained-Language-Model/tree/master/TinyBERT}} as the initializers for the corresponding student models. Those models have been distilled with one of the original BERT model's pretraining objective which is masked language model. So, they are not specific to any task and can be considered as smaller generic pretrained models.

\begin{table*}[hbt!]
\begin{center}\small
\begin{tabular}{lrrrrrrr} \toprule 
Model & BoolQ & CB & COPA & MultiRC & ReCoRD & RTE & WiC \\ \midrule
BERT$^\dagger$ (Reference, 12L, 768) & 72.7 & 80.7 & 57.0 & 41.8 & 54.9 & 65.7 & 65.6 \\ \midrule
BERT (Teacher, 12L, 768) & 75.99 & 87.96 & 64.00 & 42.00 & 64.81 & 69.68 & 72.41 \\
BERT (Student, 6L, 768) & 76.06 & 90.12 & 69.00 & 37.47 & 64.47 & 68.59 & 71.79 \\
BERT (Student, 4L, 312) & 72.63 & 87.63 & 64.00 & 36.71 & 33.68 & 64.62 & 65.20 \\ \midrule
RoBERTa (Teacher, 12L, 768) & 81.59 & 89.34 & 56.00 & 50.30 & 79.66 & 79.06 & 71.63 \\
RoBERTa (Student, 6L, 768) & 75.19 & 90.68 & 57.00 & 42.90 & 67.33 & 66.43 & 65.83 \\
\bottomrule
\end{tabular}
\end{center}
\caption{Accuracy of teacher and student models on the validation data set for each task of SuperGLUE benchmark with knowledge distillation. Model marked with $\dagger$ represents accuracy numbers on the test set provided by SustaiNLP organizers.}\label{distill_result} 
\end{table*}

\paragraph{Knowledge distillation results:}
The main goal of the knowledge distillation process is to acquire the smallest possible student model while preserving the accuracy of the teacher model. Since we are experimenting with various NLU tasks, the capacity of the optimal student model that preserves accuracy may vary  with varying level of task's difficulty. Therefore, we experiment with distilling various sized student models; then, we pick the smaller model among the distilled models that can offer higher accuracy than the original BERT model for each task. 

In our experiments, we have observed that distilled models do not work well  when distilled to a different model type. Therefore, we restricted our setup to avoid distilling RoBERTa model to BERT or vice versa. The major difference between the two model groups is the input token (sub-word) embedding. We think  that different input  embedding spaces result in different output embedding spaces, and knowledge transfer with different spaces does not work well. 

The result of the knowledge distillation on the tasks are summarized in Table \ref{distill_result} together with the teacher models' accuracy numbers on the validation data sets. It also includes the accuracy values on test data set for BERT model presented by SustaiNLP 2020 organizers\footnote{\url{https://sites.google.com/view/sustainlp2020/shared-task?authuser=0}}. We train both \textit{cased} and \textit{uncased} models using both task-specific and task-agnostic approaches, and present the model with higher accuracy values. For the more challenging tasks such as MultiRC and ReCoRD, we observe that RoBERTa based models provide better accuracy than BERT based models.

\section{Structured Pruning}
\label{sec:sp}
There has been a significant amount of research on  model's weights pruning approaches inspired by \textit{The Lottery Ticket Hypothesis} \citep{frankle2018lottery}. Furthermore, there are various papers published to apply this pruning strategy to  Transformer models including: \citep{yu2019playing, sanh2020movement, gordon2020compressing}. Most of such approaches focused on random pruning to reduce the number of parameters following the lottery ticket hypothesis. While this can reduce the size of the model on the computer storage, it may not improve the inference performance since it is not focusing on better utilization of the computing resources. Since our main focus in \textit{FastFormers} is to improve inference efficiency, randomly pruning a subset of the model's parameters may not improve performance. In this work, we focus on  structured pruning which directly reduces the computation requirements.

\citet{voita2019analyzing, michel2019sixteen, hou2020dynabert} proposed methods to prune some of the \textit{heads} of Multi-Head Attention (MHA) in Transformer architecture. \textit{DynaBERT} \citep{hou2020dynabert} additionally proposed pruning intermediate hidden states in feed-forward layer of Transformer architecture together with rewiring of these pruned attention module and feed-forward layers. In the paper, we define a target model size in terms of the number of heads and the hidden state size of feed-forward network, and use the pruning/distillation approach proposed in DynaBERT as a model compression method. This effectively reduces the dimensions of Transformer models.

\begin{figure*}[hbt!]
\begin{subfigure}{.45\textwidth}
  \includegraphics[width=1.\linewidth]{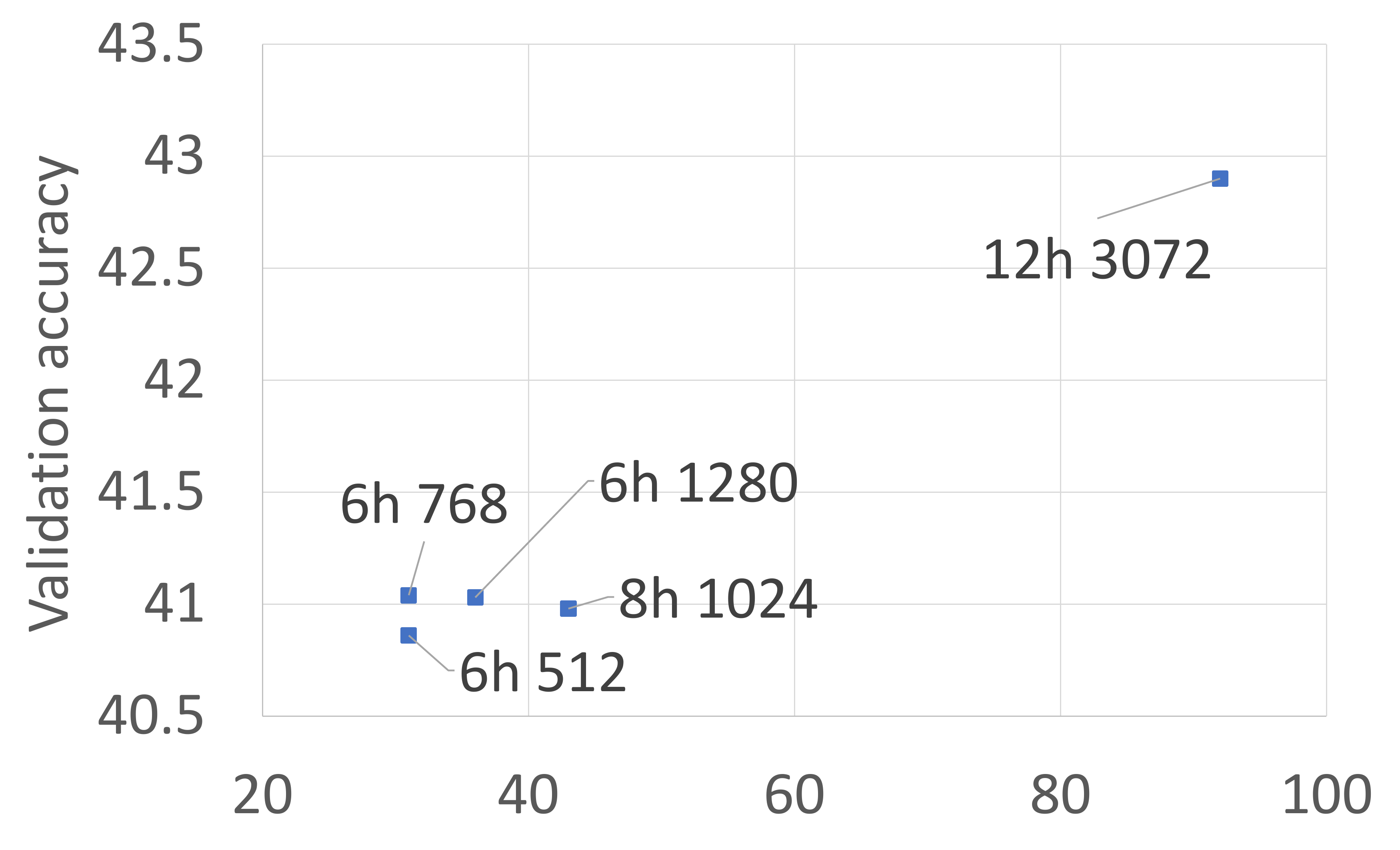}
\caption{MultiRC task}
\end{subfigure}\hfill
\begin{subfigure}{.45\textwidth}
  \includegraphics[width=1.\linewidth]{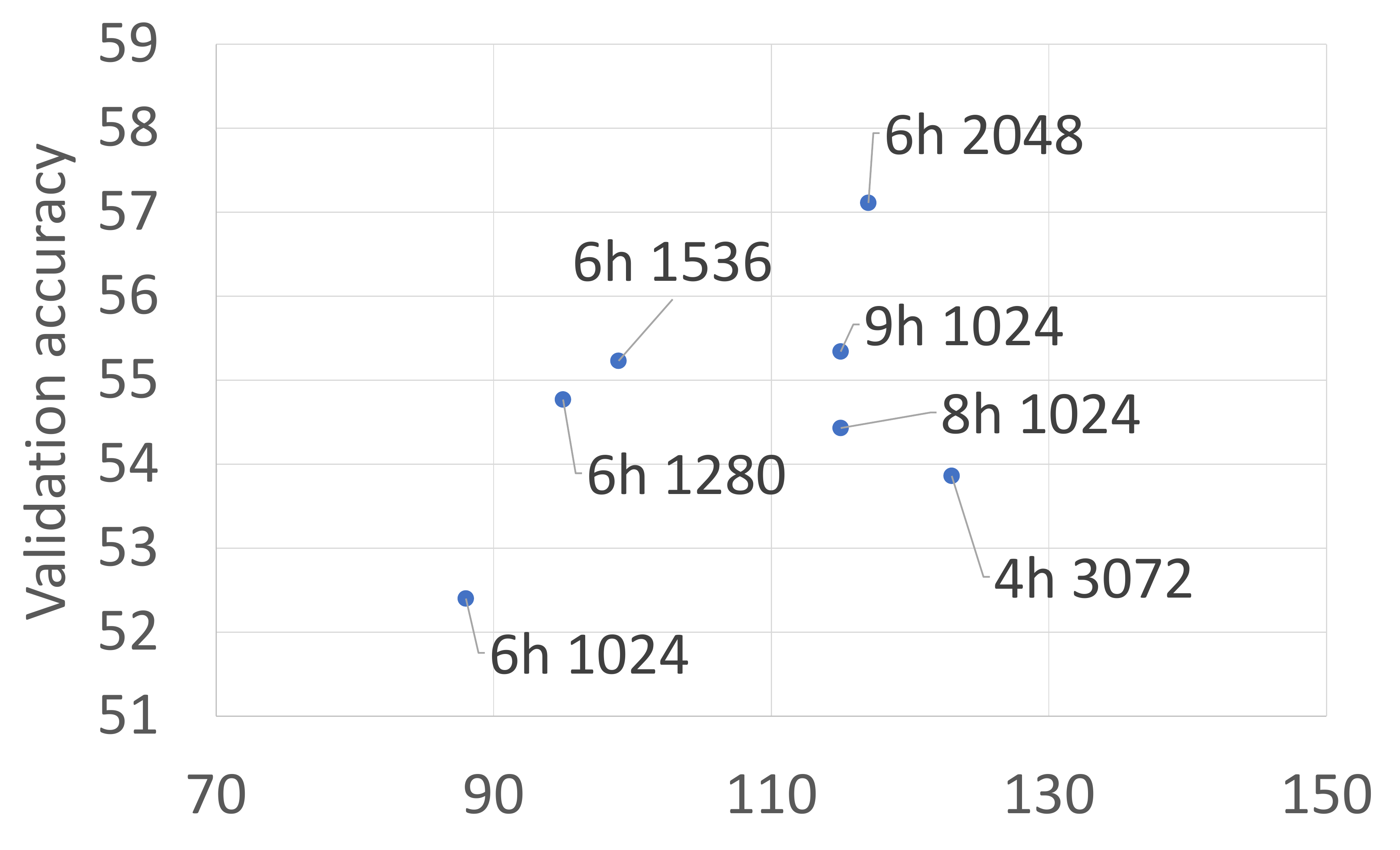}
\caption{ReCoRD task}
\end{subfigure}%
\caption{Structured pruning - inference time versus accuracy. Each data point indicates a pruned model with the number of remaining heads and the number of remaining intermediate hidden states of feed-forward layers.}
\label{fig:shrinking}
\end{figure*}

\paragraph{Structured pruning methods:}
The first step of our structured pruning method is to identify the least important \textit{heads} in MHA and the least important hidden states in the feed-forward layers. We use a first order method for computing  the importance score, which utilizes the first order gradient information proposed  by \citet{michel2019sixteen, sanh2020movement, hou2020dynabert} instead of using magnitude based pruning \citep{frankle2018lottery, gordon2020compressing}. Before doing the importance score computation, we add a mask variable to each attention head for the gradient computation of the heads. Next, we run forward and backward passes of the model on the entire validation data set, then the absolute values of the gradients are accumulated. These accumulated values are used as importance scores which we use to sort the importance of the heads and the intermediate hidden states. Based on the target model size, we select a given number of top heads and top hidden states from the network. Once the sorting and selection steps are done, we re-group and reconnect the remaining heads and hidden states which result in a smaller sized model. When heads and hidden states are pruned, we use the same pruning ratio across different layers. This enables further  optimizations to work seamlessly with  the pruned models. In our experiments, we observed that the pruned model  can get better accuracy when it goes through another round of knowledge distillation; this has also been noted in \citet{hou2020dynabert}. Therefore, we do one more knowledge distillation by using the non-pruned model as a teacher model and the pruned model as an initializer of student model.

\paragraph{Structured pruning results:}
We apply the structured pruning method mainly to MultiRC and ReCoRD tasks. For the other tasks, the test sets are not that big; therefore, knowledge distillation and  other optimizations could make the models quite efficient. In both MultiRC and ReCoRD tasks, the base model is RoBERTa based distilRoberta-base\footnote{\url{https://huggingface.co/distilRoberta-base}} which has 6 stacked layers with 768 hidden states, 12 self-attention heads and 3072 hidden states in feed-forward layer. The structured pruning method brings better efficiency by trading off the accuracy. So, it is important to pick a reasonable model size that doesn't compromise  the accuracy of the task. Figure \ref{fig:shrinking} presents trade-off between inference speed and accuracy on MultiRC and ReCoRD validation data sets. For the MultiRC task, we could get 2.97x speed-up while losing 1.9 point of accuracy by pruning 50\% of heads and 75\% of intermediate hidden states from 12 heads and 3072 hidden sizes. For the ReCoRD task, we got 1.95x speed-up while trading off 12.1 point of accuracy by pruning 50\% of heads and 50\% of hidden states in feed-froward layer. In both cases, they are still exceeding the teacher sized BERT model's accuracy.

\section{Low Precision Inference}
\label{sec:qn}
After knowledge distillation and structured pruning compression, models  can benefit form more efficient numerical computations by quantizing the model parameters on CPU and GPU \citep{rodriguez2018lower}. It has been shown that the accuracy of Transformer models doesn't get severely compromised when utilizing 8-bit or 16-bit lower precision arithmetic \citep{devlin2017sharp, kim2019research, bhandare2019efficient, zafrir2019q8bert, shen2020q, fan2020training, aji2020compressing}. Modern CPUs and GPU accelerators are capable of computing those lower numerical arithmetic efficiently. For example, Cascade Lake CPUs have a special 8-bit vector instruction set called AVX (Advanced Vector eXtensions)-512-VNNI and  V100 GPUs can utilize its efficient Tensor Cores with 16-bit floating point data.

\paragraph{8-bit quantized matrix multiplications on the CPU:}
8-bit quantized matrix multiplication brings a significant amount of speed-up compared to 32-bit floating point arithmetic, thanks to the relieved memory bandwidth bottleneck and reduced number of CPU instructions. Efficient utilization of the vector registers and CPU cache memories requires that the  parameter weight matrix  to be tiled and transposed in a cache efficient layout. This is referred as  \textit{packing} operation in the matrix multiplication. Packing itself is a non-negligible operation and repeated packing operation could cancel out all the benefits from the quantized matrix multiplications. Therefore, the result of the packing operation needs to be properly cached to get efficient performance. Moreover, some of the matrix products should stay 32-bit floating point to avoid repeated packing operations. Therefore, we do not use 8-bit matrix product for the \textit{Q}, \textit{K} inner product because both matrices are not constant. All the other matrix products have constant weight matrix, so we utilize 8-bit matrix products for them with cached weight packing. For the 8-bit quantized matrix product API, we utilize  an open source quantized matrix multiplication library FBGEMM \footnote{\url{https://github.com/pytorch/FBGEMM}} which we have integrated  into the onnxruntime framework. It provides various quantization methods and a separate matrix packing functionality explicitly. We use dynamic quantization method which decides the quantization range of input matrix dynamically every time\footnote{\url{https://pytorch.org/tutorials/recipes/recipes/dynamic_quantization.html}}. This enables the quantized values to  effectively represent all the values in the input matrix. The weight matrix' quantization range is selected for each column separately and the quantization range for the input matrix is selected for entire input tensor. In our experiments, this 8-bit quantization brings up to around 3.0x speed-up on Cascade Lake CPUs for the Transformer models by trading off small amount of accuracy loss.

\paragraph{16-bit model conversion for the GPU:}
V100 GPU supports full 16-bit operations for the Transformer architecture. Also, 16-bit floating point operations do not require special handling of inputs and outputs except for  having smaller value ranges. The impact of the numerical overflow due to the smaller range in 16-bit float points is minimal at inference time, so we have not observed any differences in accuracy. Therefore, the model can be fully converted into 16-bit floating point data type before the model is utilized for inference. This 16-bit model conversion brings quite significant speed gain, since the Transformer models are memory bandwidth bound workload. We observe up to 3.53x speed-up depending on the model settings. V100 GPU also supports 8-bit quantized arithmetic, but it is not supported with its efficient Tensor cores; hence we do not utilize 8-bit quantization on GPUs.

\section{Runtime Optimization}
\label{sec:rt}
On top of the structural and numerical optimizations applied, we can utilize  various ways to further optimize the computations. In particular, we focus on multi-processing optimizations and  computational graph optimization.

\paragraph{Multi-processing optimization:}
The evaluation machine for the SustaiNLP 2020 shared task has two Cascade Lake 6248 CPUs with 40 physical cores. The default execution engines for HuggingFace's transformers including PyTorch \footnote{\url{https://github.com/pytorch/pytorch}} and TensorFlow \footnote{\url{https://github.com/tensorflow/tensorflow}} usually use all available CPU cores for a single operator. This is not the optimal way of utilizing available CPU cores for several reasons. The operators in compressed Transformer architectures are not big enough to fully utilize the parallelism of 40 CPU cores. Therefore, the overheads of parallelizing the operation significantly overshadow the actual gains from the parallelism. Another important factor is that the parallelization to all cores reduces the cache locality and degrades the overall efficiency of CPU utilization. Therefore, we implement a multiple instance inference by leveraging multiprocessing module of python\footnote{\url{https://docs.python.org/3/library/multiprocessing.html}} programming language. It is preferable to use multi-threading instead of multi-processing to avoid additional copy of program memory, but python's multi-threading cannot really utilize multiple threads due to the Global Interpreter Lock (GIL) \footnote{\url{https://wiki.python.org/moin/GlobalInterpreterLock}}. Even with this limitation, multi-instance inference with multiprocessing brings much more efficient computation. To maximize the cache locality, each inference instance is pinned to specific physical cores using \textit{Linux}' \textit{taskset} command. Utilizing hyper-threading harms the cache utilization, so we always keep the total number of utilized threads (the number of threads per one inference instance multiplied by the number of instances) not exceeding the number of physical cores in the machine. The optimal number of multiple processes for the best efficiency varies by the model, hardware settings and the data set. We conduct experiment with all target tasks and investigate the best setting for each task. Table \ref{multiprocessing} shows one example of the speed-up achieved by optimized multi-instance inference on 1,000 samples of validation data set of ReCoRD task.

\begin{table}
\begin{center}\small
\begin{tabular}{lrr} \toprule 
Number of inference instances & Time (sec) & Speed-up \\  \midrule
Baseline (no thread control) & 433 & 1.00x \\  \midrule
1 instance (20 threads/instance) & 319 & 1.36x \\
2 instances (10 threads/instance) & 243 & 1.78x \\ 
4 instance (5 threads/instance) & 247 & 1.75x \\ 
5 instance (4 threads/instance) & 255 & 1.70x \\ 
10 instance (2 threads/instance) & 300 & 1.44x \\ 
20 instance (1 thread/instance) & 351 & 1.23x \\ 
\bottomrule
\end{tabular}
\end{center}
\caption{Speed comparison of different number of inference instances with thread control - time to perform inference on 1,000 ReCoRD validation data samples.}
\label{multiprocessing} 
\end{table}

\paragraph{Computational graph optimizations:}
Computational graph optimization can further improve the efficiency of the neural network inference by pruning unused graph nodes and fusing multiple operations together. In particular, graph node fusion reduces additional memory allocation and copy which potentially degrade the efficiency. Also, graph node fusion potentially improves parallelism by increasing the size of individual operators. We replace Gaussian Error Linear Units (GELU) with Rectified Linear Units (ReLU) for the computational efficiency while model is distilled without losing any accuracy. We fuse ReLU and bias addition operations followed by matrix multiplications into FBGEMM's fused post processing operation. Moreover, we utilize multi-head attention node fusion provided by onnxruntime. We use a customized onnxruntime based on v1.3.1.

\section{Results}
\label{sec:results}

\begin{table*}
\begin{center}\small
\begin{tabular}{lrrrrr} \toprule 
Optimization methods added & Time (sec) & Cumulative & Speed-up & Accuracy & USD for \\
 &  & speed-up &  &  & 100 M queries \\  \midrule
Baseline (PyTorch out-of-the-box, 12L, 768) & 734.35 & 1.00x & - & 74.01 & \$4,223 \\
+ dynamic sequence length & 209.29 & 3.51x & 3.51x & 74.01 & \$1,204 \\
+ knowledge distillation (4L, 312) & 22.5 & 32.64x & 9.30x & 74.04 & \$129 \\
+ 8-bit quantization + graph optimization & 9.97 & 73.66x & 2.26x & 73.43 & \$57 \\ 
+ multi-instance inference & 5.68 & 129.29x & 1.76x & 73.43 & \$33 \\  \midrule
+ structured pruning &  &  &  &  &  \\ 
\quad 25\% heads and 25\% hidden states pruned & 4.11 & 178.67x & 1.38x & 73.36 & \$24 \\ 
\quad 33\% heads and 50\% hidden states pruned & 3.14 & 233.87x & 1.81x & 72.81 & \$18 \\ 
\bottomrule
\end{tabular}
\end{center}
\caption{An ablation study of CPU inference speed-up on BoolQ validation data set with batch size 1. All performance numbers are measured on an Azure F16s-v2 instance.}\label{breakdown} 
\end{table*}

\subsection{Combined results}
\begin{figure*}
  \centering
  \includegraphics[width=1.0\linewidth]{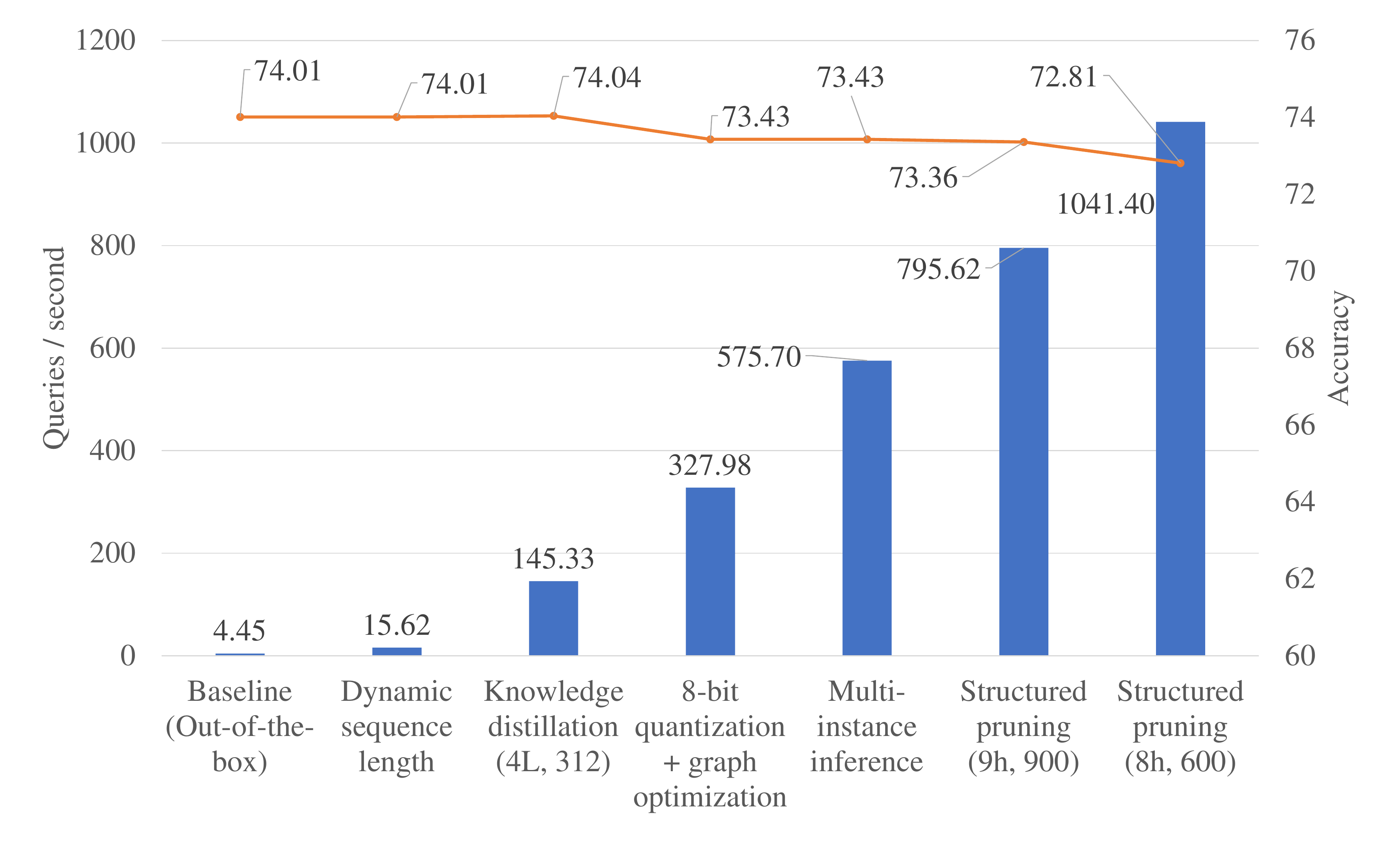}
\caption{Accuracy versus queries per second with various optimizations on CPU.}
\label{fig:ablation-graph}
\end{figure*}

Table \ref{breakdown} and Figure \ref{fig:ablation-graph} present how the proposed methods work together on BoolQ task using CPU. For the ablation study, an Azure F16s\_v2 instance which has Intel(R) Xeon(R) Platinum 8168 CPUs (8 physical cores) is utilized. When all the optimizations are applied, it could achieve around 233x speed-up while only losing 1.2 of accuracy. First, it is worth noting that the batch generation code in many frameworks including HuggingFace's transformers uses fixed sequence length for the input. Whenever there comes a shorter sentence than the fixed length for the inference, additional zeros are padded at the end of each input sentence. This degrades CPU performance greatly which has relatively small parallelism than GPU. We modify the batch generation to support dynamic sequence length for each batch to avoid such redundant computation. By doing this dynamic shape batching, we could get around 3.51x speed-up on the test CPU. On top of the dynamic shape batching, all the optimization techniques introduced are applied. As described in Section \ref{sec:kd}, knowledge distillation brings a significant speed-up which is more than 9 times faster than the original teacher model while preserving the accuracy. One thing to note is that the compressible student model size varies with the task. In this BoolQ example, 4 stacked layers with smaller hidden size (312) model could learn original teacher model's knowledge without losing any accuracy score on the validation data set. 8-bit quantization together with onnxruntime graph optimization brings around 2.26x speed-up on the 4 layer distilled model. This is lower than 3.0x speed-up acquired when the same methods is applied to 12 layer base size models mentioned in Section \ref{sec:qn}, because the system is already compressed into a smaller size. By using 8 concurrent and independent instances for inference instead of one inference instance with 8 threads, more than 1.7x additional speed-up could be achieved. Finally, by pruning heads in multi-head attentions and intermediate hidden states in feed-forward layers, another speed-up from 1.38x to 1.81x could be accomplished while trading off the accuracy. One interesting observation is that the structured pruning approach brings better speed-up when it is combined with multi-instance inference which is up to 1.81x. The same pruned model brings at most 1.26x speed-up when it is used with one inference instance. This indicates that multi-instance inference gets more performance benefits when each individual model size gets smaller.

\subsection{Shared task submissions}
We submit our optimized \textit{FastFormers} systems to SustaiNLP 2020 shared task; four systems for track 1 and two systems for track 3. For track 1, we have GPU-only submissions and hybrid submissions which utilize both CPUs and GPUs. We observe that the inference time for ReCoRD only exceeds the inference time of the other tasks all together. Therefore, for the hybrid systems, we use GPUs for ReCoRD task inference and CPUs for all the other tasks. Those CPU and GPU inferences can be executed in parallel for the best throughput. For track 3, we only use CPUs for all tasks as indicated in the shared task description. For the GPU inference, we always limit the number of GPUs we utilize to one. Our highly optimized and compressed models can be executed on a single GPU fast enough. And, the scaling of multiple GPUs is sub-linear (3.0x with 4 GPUs and 1.8x with 2 GPUs) which indicates a single GPU inference is most energy efficient. We apply different types of optimization depending on the time consumed to run the task. For most time consuming tasks,  MultiRC and ReCoRD, all compression and optimization techniques mentioned above are applied. All GPU inference uses batch size of 256 which gives the highest throughput and the best efficiency. On the other hand, a single batch works better for most of the cases on CPUs. We use batch size of 1 for all tasks except for CB (batch size of 4) and COPA (batch size of 8).

Table \ref{results} summarizes the accuracy, speed-up numbers and energy savings of \textit{FastFormers} systems prepared for the shared task. For GPUs, CB, COPA and WiC data sets are quite small, so the initial performance without any optimization took 1 or 2 seconds close to the resolution (1 second) of the wall clock time measurement. Therefore, it was hard to observe big speed improvements for those data sets. For the other tasks, we acquire a good amount of speed-up ranging from 5.8x to 12.4x by our optimization. On CPUs, model compression gives more benefits all across the board in terms of speed. The smallest gain is 9.8x for WiC data set which could not utilize onnxruntime optimization due to the control `for' loop in the output layer. This is currently not supported in onnxruntime. For the other tasks, we observe up to 40.3X speed-up. On the other hand, the energy savings while preserving BERT model accuracy measured by the shared task organizers range from 6.9X up to 125.8X. When combining all the tasks, our best systems save 22.1x energy on CPUs and 21.6x energy on GPUs.

\begin{table*}
\begin{center}\small
\begin{tabular}{lrrrrrrrr} \toprule 
Submitted systems & BoolQ & CB & COPA & MultiRC & ReCoRD & RTE & WiC & Overall \\ \midrule
BERT$^\dagger$ (Reference, 12L, 768) & 72.7 & 80.7 & 57.0 & 41.8 & 54.9 & 65.7 & 65.6 & 62.6 \\
\quad Inference time speed-up & 1.0x & 1.0x & 1.0x & 1.0x & 1.0x & 1.0x & 1.0x & 1.0x  \\
\quad Energy savings & 1.0x & 1.0x & 1.0x & 1.0x & 1.0x & 1.0x & 1.0x & 1.0x \\ \midrule \midrule
System 1 (GPU only)  & 74.0 & 82.7 & 58.0 & 41.8 & 56.2 & 66.4 & 66.0 & 63.6 \\
\quad Inference time speed-up & 5.8x & 2.0x & 2.0x & 9.1x & \textbf{12.4x} & 9.0x & 2.0x & 11.3x  \\ 
\quad Energy savings & 11.0x & 57.9x & \textbf{125.8x} & 6.9x & 23.9x & 20.5x & 28.7x & 20.3x  \\ \midrule
System 2 (GPU only)  & 74.0 & 82.7 & 58.0 & 43.2 & 56.2 & 66.4 & 66.0 & 63.8 \\
\quad Inference time speed-up & 5.8x & 2.0x & 2.0x & 9.1x & \textbf{12.4x} & 9.0x & 2.0x & 11.3x \\ 
\quad Energy savings & 11.9x & 58.3x & 80.1x & 8.1x & 25.0x & 15.9x & 22.4x & \textbf{21.6x}  \\ \midrule
System 3 (CPU/GPU hybrid) & 73.7 & 82.7 & 58.0 & 43.0 & 56.2 & 66.9 & 65.9 & 63.8 \\
\quad Inference time speed-up & \textbf{40.3x} & 22.0x & 38.0x & 25.0x & 12.4x & 22.4x & 9.8x & \textbf{14.5x} \\ 
\quad Energy savings & 13.4x & 49.4x & \textbf{92.8x} & 13.8x & 23.0x & 32.7x & 30.4x & 16.5x  \\ \midrule
System 4 (CPU/GPU hybrid) & 73.7 & 82.7 & 58.0 & 43.1 & 56.2 & 66.9 & 65.9 & 63.8\\
\quad Inference time speed-up & \textbf{40.3x} & 22.0x & 38.0x & 17.5x & 12.4x & 22.4x & 9.8x & \textbf{14.5x} \\ 
\quad Energy savings & 13.2x & 45.8x & 75.4x & 11.8x & 23.5x & 34.0x & 28.4x & 16.1x \\ \midrule
System 5 (CPU only) & 73.7 & 82.7 & 58.0 & 43.0 & 56.6 & 66.9 & 65.9 & 63.8 \\
\quad Inference time speed-up & \textbf{40.3x} & 22.0x & 38.0x & 25.0x & 15.5x & 22.4x & 9.8x & \textbf{16.6x} \\ 
\quad Energy savings & 14.0x & 41.1x & 75.0x & 15.3x & 22.5x & 34.7x & 26.2x & \textbf{22.1x} \\ \midrule
System 6 (CPU only) & 73.7 & 82.7 & 58.0 & 43.1 & 56.6 & 66.9 & 65.9 & 63.8 \\
\quad Inference time speed-up & \textbf{40.3x} & 22.0x & 38.0x & 17.5x & 15.5x & 22.4x & 9.8x & 16.1x \\ 
\quad Energy savings & 11.4x & 22.4x & 27.2x & 9.2x & 16.0x & \textbf{85.5x} & 13.5x & 15.6x \\ \midrule
\bottomrule
\end{tabular}
\end{center}
\caption{Accuracy numbers with speed-up and energy savings on the test data set of submitted systems to the SustaiNLP 2020 shared task. Model marked with $\dagger$ was accuracy numbers on the test set provided by organizers.}\label{results} 
\end{table*}

\section{Conclusion}
\label{sec:discuss}
In this paper, we have introduced \textit{FastFormers}, which achieves efficient inference-time performance for Transformer-based models on various NLU tasks. We showed that utilizing knowledge distillation, structured pruning and numerical optimization can lead to drastic improvements on inference efficiency. We showed that the improvements can be up to 200X speed-up and results in more than 200X inference  cost saving with 22X energy saving. We open source \textit{FastFormers} for the community hoping it can drive more sustainable optimizations for Transformers models. For  future directions, some other methods such as early exiting \citep{xin2020deebert, liu2020fastbert, schwartz2020right, zhou2020bert} and linear time complexity self-attention models \citep{shen2018efficient, wang2020linformer} could be added to the proposed recipes and possibly improve the efficiency further.

\clearpage
\section*{Acknowledgments}
We would like to thank the organizers of SustaiNLP 2020 for their various efforts and the reviewers for the thoughtful and helpful suggestions.

\bibliographystyle{acl_natbib}
\bibliography{emnlp2020}

\begin{thebibliography}{29}
\expandafter\ifx\csname natexlab\endcsname\relax\def\natexlab#1{#1}\fi

\bibitem[{Aji and Heafield(2020)}]{aji2020compressing}
Alham~Fikri Aji and Kenneth Heafield. 2020.
\newblock Compressing neural machine translation models with 4-bit precision.
\newblock In \emph{Proceedings of the Fourth Workshop on Neural Generation and
  Translation}, pages 35--42.

\bibitem[{Bhandare et~al.(2019)Bhandare, Sripathi, Karkada, Menon, Choi, Datta,
  and Saletore}]{bhandare2019efficient}
Aishwarya Bhandare, Vamsi Sripathi, Deepthi Karkada, Vivek Menon, Sun Choi,
  Kushal Datta, and Vikram Saletore. 2019.
\newblock Efficient 8-bit quantization of transformer neural machine language
  translation model.
\newblock \emph{arXiv preprint arXiv:1906.00532}.

\bibitem[{Devlin(2017)}]{devlin2017sharp}
Jacob Devlin. 2017.
\newblock Sharp models on dull hardware: Fast and accurate neural machine
  translation decoding on the cpu.
\newblock \emph{arXiv preprint arXiv:1705.01991}.

\bibitem[{Devlin et~al.(2018)Devlin, Chang, Lee, and
  Toutanova}]{devlin2018bert}
Jacob Devlin, Ming-Wei Chang, Kenton Lee, and Kristina Toutanova. 2018.
\newblock Bert: Pre-training of deep bidirectional transformers for language
  understanding.
\newblock \emph{arXiv preprint arXiv:1810.04805}.

\bibitem[{Fan et~al.(2020)Fan, Stock, Graham, Grave, Gribonval, J{\'e}gou, and
  Joulin}]{fan2020training}
Angela Fan, Pierre Stock, Benjamin Graham, Edouard Grave, R{\'e}mi Gribonval,
  Herv{\'e} J{\'e}gou, and Armand Joulin. 2020.
\newblock Training with quantization noise for extreme model compression.
\newblock \emph{arXiv Prepr. arXiv2004}, 7320:1--18.

\bibitem[{Frankle and Carbin(2018)}]{frankle2018lottery}
Jonathan Frankle and Michael Carbin. 2018.
\newblock The lottery ticket hypothesis: Finding sparse, trainable neural
  networks.
\newblock \emph{arXiv preprint arXiv:1803.03635}.

\bibitem[{Gordon et~al.(2020)Gordon, Duh, and Andrews}]{gordon2020compressing}
Mitchell~A Gordon, Kevin Duh, and Nicholas Andrews. 2020.
\newblock Compressing bert: Studying the effects of weight pruning on transfer
  learning.
\newblock \emph{arXiv preprint arXiv:2002.08307}.

\bibitem[{Henderson et~al.(2020)Henderson, Hu, Romoff, Brunskill, Jurafsky, and
  Pineau}]{henderson2020towards}
Peter Henderson, Jieru Hu, Joshua Romoff, Emma Brunskill, Dan Jurafsky, and
  Joelle Pineau. 2020.
\newblock Towards the systematic reporting of the energy and carbon footprints
  of machine learning.
\newblock \emph{arXiv preprint arXiv:2002.05651}.

\bibitem[{Hinton et~al.(2015)Hinton, Vinyals, and Dean}]{hinton-kd-2015}
Geoffrey Hinton, Oriol Vinyals, and Jeffrey Dean. 2015.
\newblock \href {http://arxiv.org/abs/1503.02531} {Distilling the knowledge in
  a neural network}.
\newblock In \emph{NIPS Deep Learning and Representation Learning Workshop}.

\bibitem[{Hou et~al.(2020)Hou, Shang, Jiang, and Liu}]{hou2020dynabert}
Lu~Hou, Lifeng Shang, Xin Jiang, and Qun Liu. 2020.
\newblock Dynabert: Dynamic bert with adaptive width and depth.
\newblock \emph{arXiv preprint arXiv:2004.04037}.

\bibitem[{Jiao et~al.(2019)Jiao, Yin, Shang, Jiang, Chen, Li, Wang, and
  Liu}]{jiao2019tinybert}
Xiaoqi Jiao, Yichun Yin, Lifeng Shang, Xin Jiang, Xiao Chen, Linlin Li, Fang
  Wang, and Qun Liu. 2019.
\newblock Tinybert: Distilling bert for natural language understanding.
\newblock \emph{arXiv preprint arXiv:1909.10351}.

\bibitem[{Kim et~al.(2019)Kim, Junczys-Dowmunt, Hassan, Aji, Heafield,
  Grundkiewicz, and Bogoychev}]{kim2019research}
Young~Jin Kim, Marcin Junczys-Dowmunt, Hany Hassan, Alham~Fikri Aji, Kenneth
  Heafield, Roman Grundkiewicz, and Nikolay Bogoychev. 2019.
\newblock From research to production and back: Ludicrously fast neural machine
  translation.
\newblock In \emph{Proceedings of the 3rd Workshop on Neural Generation and
  Translation}, pages 280--288.

\bibitem[{Liu et~al.(2020)Liu, Zhou, Zhao, Wang, Deng, and
  Ju}]{liu2020fastbert}
Weijie Liu, Peng Zhou, Zhe Zhao, Zhiruo Wang, Haotang Deng, and Qi~Ju. 2020.
\newblock Fastbert: a self-distilling bert with adaptive inference time.
\newblock \emph{arXiv preprint arXiv:2004.02178}.

\bibitem[{Liu et~al.(2019)Liu, Ott, Goyal, Du, Joshi, Chen, Levy, Lewis,
  Zettlemoyer, and Stoyanov}]{liu2019roberta}
Yinhan Liu, Myle Ott, Naman Goyal, Jingfei Du, Mandar Joshi, Danqi Chen, Omer
  Levy, Mike Lewis, Luke Zettlemoyer, and Veselin Stoyanov. 2019.
\newblock Roberta: A robustly optimized bert pretraining approach.
\newblock \emph{arXiv preprint arXiv:1907.11692}.

\bibitem[{Michel et~al.(2019)Michel, Levy, and Neubig}]{michel2019sixteen}
Paul Michel, Omer Levy, and Graham Neubig. 2019.
\newblock Are sixteen heads really better than one?
\newblock In \emph{Advances in Neural Information Processing Systems}, pages
  14014--14024.

\bibitem[{Rodriguez et~al.(2018)Rodriguez, Segal, Meiri, Fomenko, Kim, Shen,
  and Ziv}]{rodriguez2018lower}
Andres Rodriguez, Eden Segal, Etay Meiri, Evarist Fomenko, Y~Jim Kim, Haihao
  Shen, and Barukh Ziv. 2018.
\newblock Lower numerical precision deep learning inference and training.
\newblock \emph{Intel White Paper}, 3.

\bibitem[{Sanh et~al.(2019)Sanh, Debut, Chaumond, and
  Wolf}]{sanh2019distilbert}
Victor Sanh, Lysandre Debut, Julien Chaumond, and Thomas Wolf. 2019.
\newblock Distilbert, a distilled version of bert: smaller, faster, cheaper and
  lighter.
\newblock \emph{arXiv preprint arXiv:1910.01108}.

\bibitem[{Sanh et~al.(2020)Sanh, Wolf, and Rush}]{sanh2020movement}
Victor Sanh, Thomas Wolf, and Alexander~M Rush. 2020.
\newblock Movement pruning: Adaptive sparsity by fine-tuning.
\newblock \emph{arXiv preprint arXiv:2005.07683}.

\bibitem[{Schwartz et~al.(2020)Schwartz, Stanovsky, Swayamdipta, Dodge, and
  Smith}]{schwartz2020right}
Roy Schwartz, Gabi Stanovsky, Swabha Swayamdipta, Jesse Dodge, and Noah~A
  Smith. 2020.
\newblock The right tool for the job: Matching model and instance complexities.
\newblock \emph{arXiv preprint arXiv:2004.07453}.

\bibitem[{Shen et~al.(2020)Shen, Dong, Ye, Ma, Yao, Gholami, Mahoney, and
  Keutzer}]{shen2020q}
Sheng Shen, Zhen Dong, Jiayu Ye, Linjian Ma, Zhewei Yao, Amir Gholami,
  Michael~W Mahoney, and Kurt Keutzer. 2020.
\newblock Q-bert: Hessian based ultra low precision quantization of bert.
\newblock In \emph{AAAI}, pages 8815--8821.

\bibitem[{Shen et~al.(2018)Shen, Zhang, Zhao, Yi, and Li}]{shen2018efficient}
Zhuoran Shen, Mingyuan Zhang, Haiyu Zhao, Shuai Yi, and Hongsheng Li. 2018.
\newblock Efficient attention: Attention with linear complexities.
\newblock \emph{arXiv preprint arXiv:1812.01243}.

\bibitem[{Vaswani et~al.(2017)Vaswani, Shazeer, Parmar, Uszkoreit, Jones,
  Gomez, Kaiser, and Polosukhin}]{vaswani2017attention}
Ashish Vaswani, Noam Shazeer, Niki Parmar, Jakob Uszkoreit, Llion Jones,
  Aidan~N Gomez, {\L}ukasz Kaiser, and Illia Polosukhin. 2017.
\newblock Attention is all you need.
\newblock In \emph{Advances in neural information processing systems}, pages
  5998--6008.

\bibitem[{Voita et~al.(2019)Voita, Talbot, Moiseev, Sennrich, and
  Titov}]{voita2019analyzing}
Elena Voita, David Talbot, Fedor Moiseev, Rico Sennrich, and Ivan Titov. 2019.
\newblock Analyzing multi-head self-attention: Specialized heads do the heavy
  lifting, the rest can be pruned.
\newblock \emph{arXiv preprint arXiv:1905.09418}.

\bibitem[{Wang et~al.(2019)Wang, Pruksachatkun, Nangia, Singh, Michael, Hill,
  Levy, and Bowman}]{wang2019superglue}
Alex Wang, Yada Pruksachatkun, Nikita Nangia, Amanpreet Singh, Julian Michael,
  Felix Hill, Omer Levy, and Samuel Bowman. 2019.
\newblock Superglue: A stickier benchmark for general-purpose language
  understanding systems.
\newblock In \emph{Advances in Neural Information Processing Systems}, pages
  3266--3280.

\bibitem[{Wang et~al.(2020)Wang, Li, Khabsa, Fang, and Ma}]{wang2020linformer}
Sinong Wang, Belinda Li, Madian Khabsa, Han Fang, and Hao Ma. 2020.
\newblock Linformer: Self-attention with linear complexity.
\newblock \emph{arXiv preprint arXiv:2006.04768}.

\bibitem[{Xin et~al.(2020)Xin, Tang, Lee, Yu, and Lin}]{xin2020deebert}
Ji~Xin, Raphael Tang, Jaejun Lee, Yaoliang Yu, and Jimmy Lin. 2020.
\newblock Deebert: Dynamic early exiting for accelerating bert inference.
\newblock \emph{arXiv preprint arXiv:2004.12993}.

\bibitem[{Yu et~al.(2019)Yu, Edunov, Tian, and Morcos}]{yu2019playing}
Haonan Yu, Sergey Edunov, Yuandong Tian, and Ari~S Morcos. 2019.
\newblock Playing the lottery with rewards and multiple languages: lottery
  tickets in rl and nlp.
\newblock \emph{arXiv preprint arXiv:1906.02768}.

\bibitem[{Zafrir et~al.(2019)Zafrir, Boudoukh, Izsak, and
  Wasserblat}]{zafrir2019q8bert}
Ofir Zafrir, Guy Boudoukh, Peter Izsak, and Moshe Wasserblat. 2019.
\newblock Q8bert: Quantized 8bit bert.
\newblock \emph{arXiv preprint arXiv:1910.06188}.

\bibitem[{Zhou et~al.(2020)Zhou, Xu, Ge, McAuley, Xu, and Wei}]{zhou2020bert}
Wangchunshu Zhou, Canwen Xu, Tao Ge, Julian McAuley, Ke~Xu, and Furu Wei. 2020.
\newblock Bert loses patience: Fast and robust inference with early exit.
\newblock \emph{arXiv preprint arXiv:2006.04152}.

\end{thebibliography}

\end{document}